\documentclass{article}
\usepackage[dvipsnames, svgnames, x11names]{xcolor}
\usepackage[pagebackref=true, 
            breaklinks=true, 
            letterpaper=true, 
            colorlinks,
            citecolor=ForestGreen, 
            linkcolor=red, 
            bookmarks=false,
            urlcolor=violet]{hyperref}
\usepackage{spconf,amsmath,graphicx}
\usepackage{booktabs}
\usepackage{amsfonts,amssymb}
\usepackage{bbding}
\usepackage{bbm}

\usepackage{algpseudocode}
\usepackage{algorithmicx,algorithm}

\title{Prototype Knowledge Distillation for Medical Segmentation with Missing Modality}

\name{Shuai Wang$^1$, Zipei Yan$^2$, Daoan Zhang$^3$, Haining Wei$^1$, Zhongsen Li$^1$, Rui Li$^1$}
\address{$^1$Tsinghua University, $^2$The Hong Kong Polytechnic University, \\
$^3$Southern University of Science and Technology}
\begin{document}
%
\maketitle
\begin{abstract}
Multi-modality medical imaging is crucial in clinical treatment as it can provide complementary information for medical image segmentation. 
However, collecting multi-modal data in clinical is difficult due to the limitation of the scan time and other clinical situations. 
As such, it is clinically meaningful to develop an image segmentation paradigm to handle this missing modality problem.
In this paper, we propose a prototype knowledge distillation (ProtoKD) method to tackle the challenging problem, especially for the toughest scenario when only single modal data can be accessed.
Specifically, our ProtoKD can not only distillate the pixel-wise knowledge of multi-modality data to single-modality data
but also transfer intra-class and inter-class feature variations, such that the student model could learn more robust feature representation from the teacher model and inference with only one single modality data. Our method achieves state-of-the-art performance on BraTS benchmark. The code is available at \url{https://github.com/SakurajimaMaiii/ProtoKD}.
\end{abstract}
%


\section{Introduction}
 Multi-modality imaging is significant in the medical image analysis field, as it provides complementary information for medical diagnosis \cite{Menze2015,Maier2017,wu2022gamma}. Although multi-modality imaging usually produces accurate diagnosis, it is often difficult to collect a complete set of multi-modality images due to data corruption or various scanning protocols in the clinical scenario. Consequently, a robust medical image segmentation method is highly desired to tackle the missing modality problem.

Three main streams of approaches have been proposed to tackle this challenging problem where there are missing modalities at inference time. The first stream is to synthesize missing modalities to complete the test set \cite{ShenZWXPTHSMCWX21,0005KMY19}, which requires training a generative model to generate missing modalities. These methods usually require extra training and are hard to complete various modalities when only one modality is available at inference time. The second stream aims to learn a shared latent space which includes modality invariant information among accessible domains \cite{vanTulder2019,hved,ZhouCVR21,ChenDJCQH19,ZhangHYLWHZHZ22}. These strategies achieve good performance but they achieve bad results when only a single modality is available. Recently, some knowledge distillation \cite{hinton_kd,Lopez-PazBSV15}, based methods have been proposed to tackle the challenging problem that there is only one modality available at inference time \cite{hu2020knowledge,pmkl}. They aim to transfer knowledge from the teacher model trained using multi-modality images to the student model that is only trained with one modality.

\begin{figure*}[t]
    \centering
    \includegraphics[width=\textwidth]{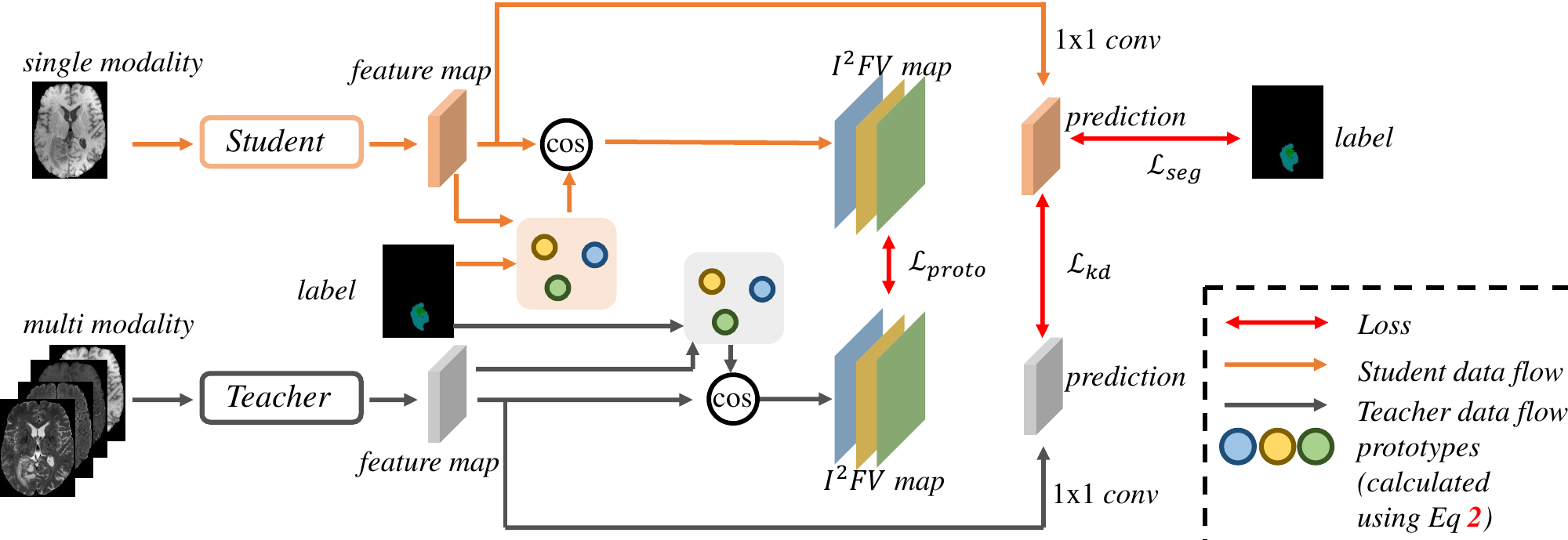}
    \vspace{-0.8cm}
    \caption{The overview of the proposed method. Both the teacher and student models share the same architecture except for different inputs.}
    \label{fig:framework}
    \vspace{-1em}
\end{figure*}
Common knowledge distillation-based approach to tackle the missing modality problem is to directly align the output features of the student and teacher models \cite{hinton_kd,hu2020knowledge,pmkl,RomeroBKCGB14}. However, medical images are structural and analogous to each other, and image segmentation task requires detailed structure semantic information for pixel classification \cite{zhang2022rethinking}. Thus for medical image segmentation, the relations among intermediate features in the teacher model should be considered and inherited by the student model.

Motivated by this, we propose a prototype knowledge distillation (ProtoKD) by matching \textbf{I}ntra-class and \textbf{I}nter-class \textbf{F}eature \textbf{V}ariation ($I^2FV$) between the student model and the teacher model for medical image semantic segmentation. Our method takes the regional information in medical images into account to benefit the segmentation result.

As illustrated in Figure \ref{fig:framework}, We first compute prototypes for every class, then generate the proposed $I^2FV$ map by calculating the inter- and intra-relations between pixel features and prototypes. After that, we transfer the knowledge in the dense similarity maps from the teacher model to the student model. Intuitively, the teacher model is trained with multi-modality images and usually produces better and more integral feature presentations than the student model. We evaluate our method on BraTS~\cite{Menze2015} benchmark. The experiment results show that our method not only consistently improves unimodal segmentation baseline but also achieves a new state-of-the-art performance.
\begin{algorithm}[t]
\caption{Prototype Knowledge Distillation}
\label{alg:alg}
\begin{algorithmic}
    \Require teacher model with parameter $\theta_t$ including backbone $f_t$ and classification head $h_t$; \\
            student model with parameter $\theta_s$ including backbone $f_s$ and classification head $h_s$; \\
                 single/multi-modality input $x$/$x^*$, ground truth $y$; \\
                 iteration numbers $N$, learning rate $\eta$.
    \Ensure $\theta_s$
    \State Initialize $\theta_s$ and $\theta_t$ randomly 
    \For {$i=1:N$}  \algorithmiccomment{Pre-train teacher model}
        \State $p^t=h_t(f_t(x^*))$
        \State $\mathcal{L}_{seg} = \ell_{ce}(p^t,y) + \ell_{dice}(p^t,y)$ \algorithmiccomment{Eq \ref{eq:loss_seg}}
        \State $\theta_t \leftarrow \theta_t - \eta\nabla_{\theta_t}\mathcal{L}_{seg}$ \algorithmiccomment{Update teacher model}
    \EndFor
    \For {$i=1:N$}  \algorithmiccomment{Train student model}
        \State $z^s=f_s(x),p^s=h_s(z^s)$
        \State $z^t=f_t(x^*),p^t = h_t(z^t)$
        \State $\mathcal{L}_{seg} = \ell_{ce}(p^s,y) + \ell_{dice}(p^s,y)$ \algorithmiccomment{Eq \ref{eq:loss_seg}}
        \State $\mathcal{L}_{kd} = \text{KL}\left( \sigma \left( p^s/T\right)|| \sigma\left( p^t/T \right) \right)$  \algorithmiccomment{Eq \ref{eq:loss_kd}}
        \State Calculate $\mathcal{L}_{proto}$ according to Eq \ref{eq:loss_proto}
        \State $\mathcal{L} = \mathcal{L}_{seg} + \alpha \mathcal{L}_{kd} + \beta \mathcal{L}_{proto}$ \algorithmiccomment{Objective function}
        \State $\theta_s \leftarrow \theta_s - \eta\nabla_{\theta_s}\mathcal{L}$ \algorithmiccomment{Update student model}
    \EndFor \\
    \Return $\theta_s$
\end{algorithmic}
\end{algorithm}
\section{Method}
The overview of our method is illustrated in Figure \ref{fig:framework}. Our framework aims to transfer the knowledge from the well-trained teacher model to the student model, where the teacher model takes the multi-modality input, while the student model only takes the single-modality input. Such that the student can make robust predictions as well as the teacher model, by only referring to the single-modality inputs. Except for different inputs, both the teacher and student model share the same architecture. In general, we first obtain a well-trained teacher model by training it using multi-modality data. Then, we transfer the knowledge from the teacher to the student model in a knowledge-distillation manner. The details of our method are presented in the following sections.
\subsection{Pixel-wise Knowledge Distillation}
We follow the common knowledge distillation approach proposed in \cite{hinton_kd} because the segmentation problem could be formulated as the pixel-level classification problem. We encourage the student model to learn knowledge from the teacher model by minimizing the Kullback-Leibler divergence between the prediction from the student model and the teacher model. The pixel-wise knowledge distillation loss is formulated as follows
\begin{equation}
    \mathcal{L}_{kd}(p^s,p^t) = \text{KL}\left( \sigma \left( p^s/T\right)|| \sigma\left( p^t/T \right) \right),
    \label{eq:loss_kd}
\end{equation}
where $\sigma$ denotes softmax operation. KL denotes Kullback-Leibler divergence and $T$ is the temperature hyper-parameter. We empirically set $T=10$. $p^s$ and $p^t$ denote the prediction of the student model and teacher model, respectively.
\subsection{Prototype Knowledge Distillation}
Although pixel-wise distillation encourages similar feature distributions per pixel between the prediction from the student and teacher model, the inner semantic correlations among the whole distribution are not fully exploited. 

Motivated by this, we proposed to consider the correlation of intra- and inter-class feature variation, such that inner semantic correlations are explicitly exploited. We accomplish this goal by transferring the knowledge from the well-trained teacher model to the student model. The intuition behind this is that the teacher model can capture more robust intra- and inter-class feature representation as it is trained with multi-modality data. In our method, the correlation of intra- and inter-class feature representation can be captured by the similarity between features of \textit{all} pixels and prototypes of \textit{all} classes.

Prototype learning is widely used in the few-shot learning field \cite{SnellSZ17}, which represents the embedding center of every class. In our method, for class $k$, prototype $c_k$ is formulated as follows
\begin{equation}
    c_k = \frac{\sum_{i}z_i \mathbbm{1}[y_i=k] }{\sum_{i}\mathbbm{1}[y_i=k]},
\end{equation}
where $z_i$ is the feature embedding of pixel $i$ and $y_i$ denotes the ground truth of pixel $i$. $\mathbbm{1}$ is an indicator function, outputting value 1 if the argument is true or 0 otherwise.

After that, we define inter- and intra-class feature variation ($I^2FV$) similarity of pixel $i$ as
\begin{equation}
    M_k(i) = \frac{z_i^T c_k}{\Vert z_i \Vert  \Vert c_k \Vert},
\end{equation}
where $M_k(i)$ denotes similarity between feature of pixel $i$ and prototype $c_k$ and $\Vert a \Vert$ represents $\ell_2$ norm of vector $a$. If pixel $i$ belongs to class $k$, $M_k(i)$ represent intra-class feature variation. If pixel $i$ does not belong to class $k$, $M_k(i)$ could represent inter-class feature variation. As shown in Figure \ref{fig:framework}, both the teacher and student models generate their $I^2FV$ maps, respectively.  

As we aim to transfer $I^2FV$ map from the teacher model to the student model, we use $L_2$ distance as the objective function to minimize the distance of two $I^2FV$ maps. Then, the prototype knowledge distillation loss is formulated as follows
\begin{equation}
    \mathcal{L}_{proto} =\frac{1}{\left| \mathcal{N} \right|K} \sum_{i\in \mathcal{N}} \sum_{k=1}^{K} \Vert M_k^{s}(i) - M_k^{t}(i) \Vert ^2,
    \label{eq:loss_proto}
\end{equation}
where $M_k^{s}$ and $M_k^{t}$ denote $I^2FV$ similarity map of student model and teacher model, respectively.

For the medical image segmentation task, hybrid segmentation loss combining cross entropy loss and Dice loss \cite{vnet} is widely used, which is formulated as follows
\begin{equation}
    \mathcal{L}_{seg}(p,y) = \ell_{ce}(p,y) + \ell_{dice}(p,y),
    \label{eq:loss_seg}
\end{equation}
where $y$ denotes ground truth. $\ell_{ce}$ denotes standard cross entropy loss and $\ell_{dice}$ denotes Dice loss \cite{vnet}.

Finally, the final objective function consists of segmentation loss (Eq \ref{eq:loss_seg}), pixel-wise knowledge distillation loss (Eq \ref{eq:loss_kd}) and prototype knowledge distillation loss (Eq \ref{eq:loss_proto}):
\begin{equation}
    \mathcal{L} = \mathcal{L}_{seg} + \alpha \mathcal{L}_{kd} + \beta \mathcal{L}_{proto},
    \label{eq:loss}
\end{equation}
where $\alpha$ and $\beta$ are hyper-parameters to balance the loss components \footnote{We set $\alpha=10$ and $\beta=0.1$ for all experiments.}. We summarize our method in Algorithm \ref{alg:alg}.

\section{Experiments}
\subsection{Setup}
\textbf{Dataset}. We evaluate our method on the BraTS 2018 Challenge dataset \cite{Menze2015}, which contains 285 cases with manually annotated labels. Each subject has four MRI modalities, including T1, T2, T1ce and Flair. Annotation is manually performed by radiologists, which includes enhancing tumor (ET), edema (ED) and non-enhancing tumor core (NET). For pre-processing, each volume is normalized to zero mean and unit variance. We randomly crop each volume to $96\times128\times128$ to feed the network due to limited GPU memory. We randomly split 285 cases into train(70\%)/validation(10\%)/test(20\%), respectively.

\textbf{Baselines.} We first implement a \textbf{Unimodal} baseline, which is trained in a supervised manner using only one modality. Furthermore, we compare our method with (1) \textbf{U-HVED} \cite{hved}, a representation learning method that embeds different modalities to a shared latent space, (2) \textbf{KD-Net} \cite{hu2020knowledge} and (3) \textbf{PMKL} \cite{pmkl}, two approaches using knowledge distillation. PMKL \cite{pmkl} is implemented with the same network as our method (i.e., VNet) using public released code \footnote{\url{https://github.com/cchen-cc/PMKL}}.

\textbf{Evaluation Metric}. Our task is to segment each subject into three regions including whole tumor (WT), tumor core (CO) and enhancing core (EC). We evaluate the performance using Dice Score (DSC), which is commonly used in medical image analysis  and is defined as
\begin{equation}
    \text{Dice}(P,G) = \frac{2\times|P\cap G|}{|P|+|G|},
\end{equation}
where $P$ denotes outputs of the model and $G$ denotes ground truth. DSC measures overlap between prediction and ground truth, and higher DSC indicates better performance.

\textbf{Implementation Details}. We use VNet~\cite{vnet} as our segmentation backbone\footnote{Note that our method is model-agnostic, which could be adopted by different segmentation backbones}. We first train the Teacher model with 1000 epochs using four modality data according to $\mathcal{L}_{seg}$ (Eq \ref{eq:loss_seg}). Specifically, we set the batch size to 4 and use Adam optimizer with learning rate $\eta=1e^{-3}$ and weight decay equals $1e^{-5}$. Besides, the learning rate $\eta$ is reduced by multiplying with $(1-\textrm{epoch}/ \textrm{max}\_\textrm{epoch})^{0.9}$ during the training. After that, we fix the Teacher model and then train the student model using the proposed Prototype Knowledge Distillation for 1000 epochs. We perform model selection on the validation set with reference to the highest DSC.
\begin{table*}[t]
  \centering
  \caption{Results on BraTS. Metric: Dice Score (DSC). The best result in each modality is \textbf{bold-faced}. The results in the first row (Teacher) are trained with full modality images. And * denotes statistical significance in paired t-test (* indicates $p \leq 0.05$).}
  \resizebox{\textwidth}{!}{
    \begin{tabular}{l|llll|llll|llll|llll}
    \toprule
          & \multicolumn{4}{c|}{T1}       & \multicolumn{4}{c|}{T2}       & \multicolumn{4}{c|}{T1ce}     & \multicolumn{4}{c}{Flair} \\
\cmidrule{2-17}          & WT    & CO    & EC    & Avg   & WT    & CO    & EC    & Avg   & WT    & CO    & EC    & Avg   & WT    & CO    & EC    & Avg \\
    \midrule
    Teacher & 86.26 & 79.10  & 77.44 & 80.93 & -     & -     & -     & -     & -     & -     & -     & -     & -     & -     & -     & -  \\
    \midrule
    Unimodal & 72.96  & 65.59  & 37.77  & 58.77  & 82.65  & 66.76  & 45.32  & 64.91  & 71.41   & 73.30  & \textbf{76.36}  & 73.69   & 81.91        & 63.57  & 40.74  & 62.07 \\
    U-HVED \cite{hved}   & 52.40  & 37.20  & 13.70  & 34.43  & 80.90  & 54.10  & 30.80  & 55.27  & 62.40   & 66.70  & 65.50  & 64.87   & 82.10        & 50.40  & 24.80  & 52.43 \\
    KD-Net \cite{hu2020knowledge}   & \textbf{79.62}  & 59.83  & 33.69  & 57.72  & \textbf{85.74}  & 66.79  & 33.63  & 62.05  & \textbf{78.87}   & 80.83  & 70.52  & 76.74*  & \textbf{88.28}        & 64.37  & 33.39  & 62.01 \\
    PMKL \cite{pmkl}     & 71.31  & 64.26  & 41.37  & 58.98  & 81.00  & 67.92  & 47.09  & 65.34* & 70.50   & 76.92  & 75.54  & 74.32   & 84.11        & 62.21  & 41.35  & 62.56 \\
    ProtoKD (Ours)  & 74.46  & \textbf{67.34}   & \textbf{47.41}  & \textbf{63.07}* & 81.83  & \textbf{68.29}  & \textbf{47.35}  & \textbf{65.82}* & 74.67  & \textbf{81.48}  & 76.01  & \textbf{77.39}*  & 84.64        & \textbf{65.56}  & \textbf{42.30}  & \textbf{64.17}* \\
    \bottomrule
    \end{tabular}
    }
    \vspace{-1.5em}
  \label{tab:main_results}
\end{table*}%

\begin{figure}[t]
    \centering
    \includegraphics[width=\linewidth]{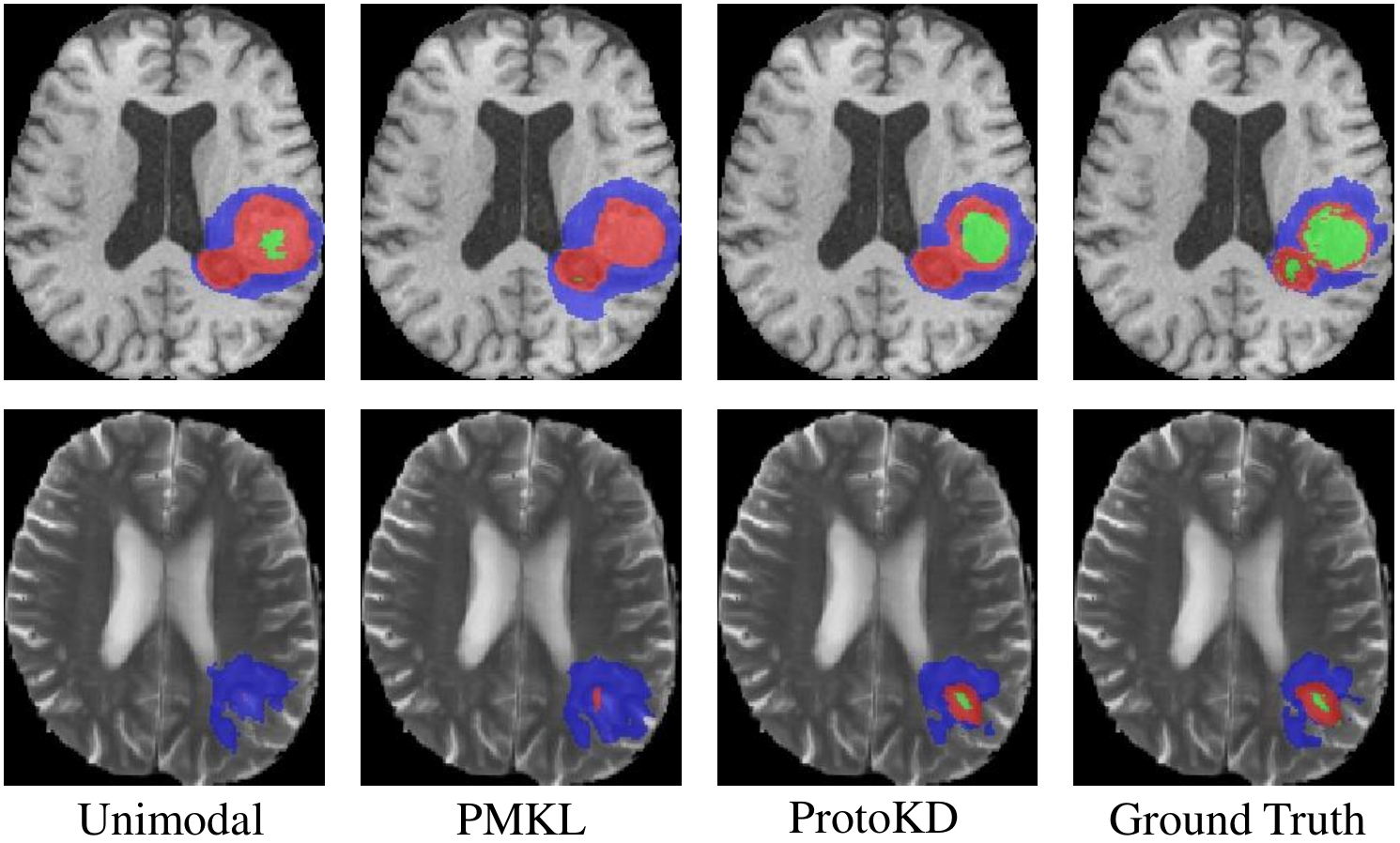}
    \vspace{-0.8cm}
    \caption{Visualization of predictions from different methods on BraTS samples. The enhancing tumor, edema and non-enhancing tumor core are marked in \textcolor{red}{red}, \textcolor{ForestGreen}{green} and \textcolor{blue}{blue} color, respectively.}
    \label{fig:visualization}
    \vspace{-1.5em}
\end{figure}

\begin{table}[t]
  \centering
  \caption{Ablation study of different components on T1 modality. Metric: DSC(\%\textuparrow).}
    \begin{tabular}{ccc|c}
    \toprule
    $\mathcal{L}_{seg}$ & $\mathcal{L}_{kd}$ & $\mathcal{L}_{proto}$ & DSC(\%\textuparrow) \\
    \midrule
   \Checkmark     &       &       & 58.77  \\
    \Checkmark     & \Checkmark     &       & 60.43(+1.66\textuparrow) \\
    \Checkmark     &       & \Checkmark     & 61.70(+2.93\textuparrow)\\
    \Checkmark     & \Checkmark     & \Checkmark     & 63.07(+4.30\textuparrow) \\
    \bottomrule
    \end{tabular}%
  \label{tab:ablation_stduy}
  \vspace{-1.5em}
\end{table}%

\begin{table}[t]
  \centering
  \caption{Ablation study on transferring knowledge of intra-class and inter-class feature variation. Metric: DSC(\%\textuparrow).}
    \begin{tabular}{cc|cccc}
    \toprule
    \multicolumn{2}{c|}{Feature variation} & \multicolumn{4}{c}{Modality} \\
    \midrule
    Intra-class & Inter-class & T1 & T2 & T1ce & Flair \\
    \midrule
    \Checkmark & &    62.50 &  65.40  &  76.80 & 63.10 \\
    \Checkmark & \Checkmark &    63.07 & 65.82 & 77.39 & 64.17 \\
    \bottomrule
    \end{tabular}
  \label{tab:intra_inter}
  \vspace{-1.5em}
\end{table}
\subsection{Results}

\textbf{Quantitative Results.} We report the segmentation results for quantitative comparison in Table~\ref{tab:main_results}. First, our method ProtoKD generally improves unimodal baseline. For different modalities, our method increases DSC by 4.3\%, 0.9\%, 3.7\% and 2.1\%, respectively. Furthermore, our method performs better than compared methods, such as PMKL \cite{pmkl} and KD-Net \cite{hu2020knowledge}. This endorses the benefits of transferring inter-class and intra-class feature variation, which provides better feature representation for the student model. Furthermore, we conduct the paired t-test between different methods and unimodal baseline to analyze whether the performance gain of different methods is statistically significant. As shown in Table \ref{tab:main_results}, the improvement of our method is statistically significant for all modalities.

\textbf{Qualitative Results.} We present the qualitative results  in Figure~\ref{fig:visualization}. As shown in Figure~\ref{fig:visualization}, we can observe that our method produces more compact shapes and is more similar to ground truth compared with other methods.
\vspace{-1em}
\subsection{Ablation Study}
To get a better understanding of the effectiveness of key components in our proposed method, we conduct two ablation studies. 

First, we study the effectiveness of different components in the object function, i.e., pixel-wise knowledge distillation $\mathcal{L}_{kd}$ and prototype knowledge distillation $\mathcal{L}_{proto}$. The results are reported in Table~\ref{tab:ablation_stduy}. Based on the vanilla segmentation loss $\mathcal{L}_{seg}$, both pixel-wise knowledge distillation $\mathcal{L}_{kd}$ and prototype knowledge distillation $\mathcal{L}_{proto}$ can improve the performance with respect to 1.66\% and 2.93\%, separately. And our proposed prototype knowledge distillation outperforms pixel-wise knowledge distillation. Besides, the best results are produced by the combination of vanilla segmentation loss, pixel-wise knowledge distillation and prototype knowledge distillation, which further illustrates the compatibility of our proposed prototype knowledge distillation.

Furthermore, we study the importance and effectiveness of learning inter-class feature variation. The results are reported in Table~\ref{tab:intra_inter}. As shown in Table~\ref{tab:intra_inter}, we can observe that with additional transfer knowledge of inter-class feature variation, the performance generally improves compared to only intra-class feature variation. 
\section{Conclusion}
In this paper, we propose a novel knowledge distillation-based method to tackle the missing modality problem in medical image segmentation. We introduce intra- and inter-class feature variation distillation to alleviate the difference in feature distribution between the student model and teacher model. This method facilitates the student model to capture more robust features by transferring knowledge from the teacher model that the teacher usually has better feature representation. We conduct extensive experiments on BraTS 2018 benchmark and experimental results demonstrate the effectiveness of our method.

\noindent
\textbf{Acknowledgement}. This work was supported by the National Natural Science Foundation of China, 81971604 and the Grant from the Tsinghua Precision Medicine Foundation, 10001020104.
\bibliographystyle{IEEEbib}
\bibliography{refs}
\end{document}